\documentclass[letterpaper, 10 pt, journal, twoside]{IEEEtran}
\usepackage{cite}%make refs turn up in ordeFr

\usepackage{amsmath,amssymb,amsfonts}

\usepackage{amsthm}
\usepackage{mathtools}
\usepackage{siunitx}

\usepackage{thmtools,thm-restate}
\newtheorem{assumption}{Assumption}
\newtheorem{definition}{Definition}
\newtheorem{theorem}{Theorem}

\usepackage{algorithm, algorithmic}
\usepackage{booktabs,caption}

\usepackage{tabu}
\usepackage{tabularx}
\newcolumntype{Y}{>{\centering\arraybackslash}X}
\usepackage{multirow}
\usepackage{mwe}
\usepackage{caption}

\usepackage{subfig}
\usepackage{float}
\usepackage{xcolor}
\usepackage{graphicx}
\usepackage{wrapfig}
\usepackage{textcomp}
\usepackage{lipsum}
\usepackage{cuted}
\usepackage{tikz}
\newcommand*\circled[1]{\tikz[baseline=(char.base)]{
  \node[shape=circle,draw,inner sep=0pt, minimum size=4mm] (char) {#1};}}
\newcommand{\realR}{\mathbb{R}}

\makeatletter
 \let\NAT@parse\undefined
 \makeatother
\usepackage[pdftex,
colorlinks=true,
bookmarks=true,
citecolor=red,
urlcolor=black,
linkcolor=blue,
pagebackref=false,
pdfstartview=Fit,
breaklinks=true
]{hyperref}

\begin{document}

\title{\LARGE \bf
Hierarchical Primitive Composition: Simultaneous Activation of Skills with Inconsistent Action Dimensions in Multiple Hierarchies}

\author{Jeong-Hoon~Lee~and~Jongeun~Choi$^{*}$% <-this % stops a spaced

\thanks{Manuscript received: February, 24\textsuperscript{th}, 2022; Revised May, 11\textsuperscript{th}, 2022; Accepted May, 31\textsuperscript{st}, 2022.}%Use only for final RAL version
\thanks{This paper was recommended for publication by Editor Jens Kober upon evaluation of the Associate Editor and Reviewers' comments.}%
\thanks{The authors are with the School of Mechanical Engineering, Yonsei University, Seoul, 03722, Republic of Korea.}%
\thanks{This work was supported by the National Research Foundation of Korea (NRF) funded by the Korea Government (MSIT), under Grant 2021R1A2B5B01002620. (Corresponding author: Jongeun Choi.)}%
\thanks{Digital Object Identifier (DOI): see top of this page.}
}

\markboth{IEEE Robotics and Automation Letters. Preprint Version. Accepted June, 2022}
{Lee \MakeLowercase{\textit{et al.}}: HPC} 

\maketitle
% \thispagestyle{empty}
% \pagestyle{empty}

%%%%%%%%%%%%%%%%%%%%%%%%%%%%%%%%%%%%%%%%%%%%%%%%%%%%%%%%%%%%%%%%%%%%%%%%%%%%%%%%
\begin{abstract}
    Deep reinforcement learning has shown its effectiveness in various applications, providing a promising direction for solving tasks with high complexity. However, naively applying classical RL for learning a complex long-horizon task with a single control policy is inefficient. Thus, policy modularization tackles this problem by learning a set of modules that are mapped to primitives and properly orchestrating them. In this study, we further expand the discussion by incorporating simultaneous activation of the skills and structuring them into multiple hierarchies in a recursive fashion. Moreover, we sought to devise an algorithm that can properly orchestrate the skills with different action spaces via multiplicative Gaussian distributions, which highly increases the reusability. By exploiting the modularity, interpretability can also be achieved by observing the modules that are used in the new task if each of the skills is known. We demonstrate how the proposed scheme can be employed in practice by solving a pick and place task with a 6 DoF manipulator, and examine the effects of each property from ablation studies.
\end{abstract}
%%%%%%%%%%%%%%%%%%%%%%%%%%%%%%%%%%%%%%%%%%%%%%%%%%%%%%%%%%%%%%%%%%%%%%%%%%%%%%%%

\section{Introduction}
\label{Introduction}
Recent studies on deep reinforcement learning (RL) have proven their effectiveness when applied to the field of robotics within simulators~\cite{kilinc2019reinforcement} or in the real world~\cite{hwangbo2017control, lee2020learning, kim2020unexpected, kim2021vision, lim2020prediction}. However, it is inefficient to naively apply classical RL for learning a complex long-horizon task with a single control policy (flat policy), as it is required for the agent to explore a vast state-action space and collect tens to hundreds of millions of samples till it sufficiently encounters a proper learning signal~\cite{buckman2018sample, yang2020multi,li2020multi}. Moreover, it has been reported that such a method is susceptible to overfitting the policy to idiosyncrasies of the environment~\cite{zhang2018dissection}.

A more practical approach is to reduce the scope of the exploration by reusing the prior knowledge obtained from solving relevant and comparatively easier tasks or \textit{subtasks}. Among various methods of leveraging prior knowledge, the transfer learning scheme proposes to compose policies that are trained from previous tasks and fine-tune on newly appointed tasks~\cite{devin2017learning, scheiderer2020transfer}. Although this approach had successfully reduced the amount of data required, knowledge transfer via fine-tuning incurs catastrophic forgetting when the gradients of the new task flow through~\cite{rusu2016progressive}.

\begin{figure}[t]
    \centering
    \includegraphics[width=\columnwidth]{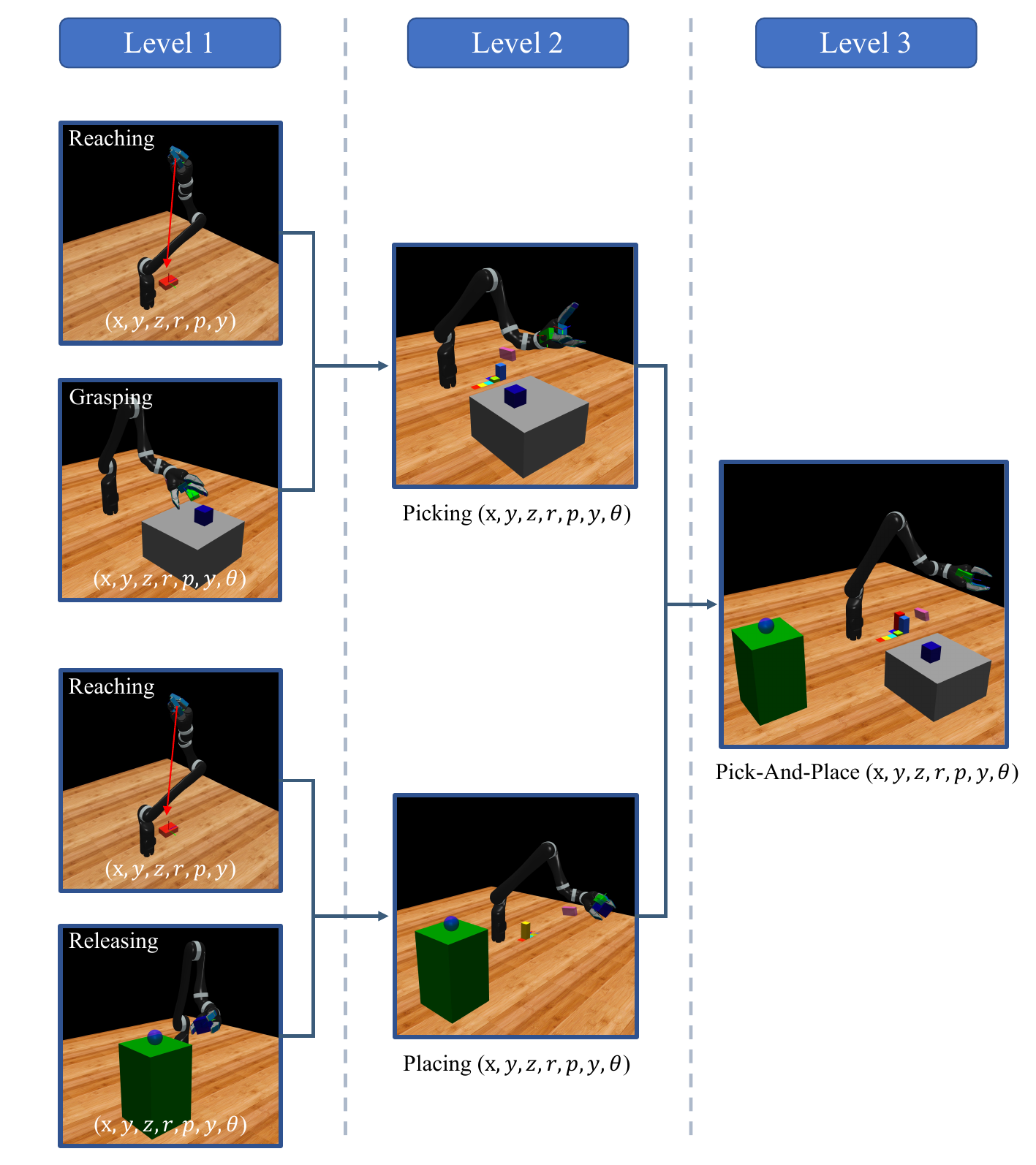}
    \caption{Schematics of the \texttt{pick-and-place} task formulation. The compound \texttt{pick-and-place} is assigned to level 3. It can be decomposed into level 2 tasks, \texttt{picking} and \texttt{placing}. These tasks can again be decomposed into level 1 tasks, \{\texttt{reaching}, \texttt{grasping}\} and \{\texttt{reaching}, \texttt{releasing}\}, respectively. 
    % While \texttt{releasing} has \blue 6 action dimensions (incremental pose of the end-effector), \black others include gripper angle as an additional componenet of the controllable action.
    }
    \label{fig:task_schematic}
\end{figure}

Policy modularization, on the other hand, suggests learning a set of modules that are mapped to primitive \textit{skills} and properly orchestrating them to tackle such problems~\cite{devin2017learning, andreas2017modular, das2018neural,xu2019toward, beyret2019dot, lee2019learning, peng2019mcp,hasenclever2020comic,pore2020simple,lee2021adversarial,dalal2021accelerating}. If the modules are fixed, not only it prevents forgetting, but it also offers interpretability by observing the modules that are used in the new task~\cite{alet2018modular}. A straightforward way of coordinating these skills is to arrange them to be activated one at a particular time. This, however, will restrict the behavior of the agent to the prescribed primitive while it is necessary to perform more than one subtask simultaneously as the complexity of the system grows~\cite{peng2019mcp}. Thus, \cite{peng2019mcp,xu2019toward} studied concurrent activation of skills to mitigate such a problem, but they required every skill to have identical action dimensions. Lastly, the task should be divided into multiple hierarchies. For instance, even a simple robotic task such as pick-and-place, as shown in Fig.~\ref{fig:task_schematic}, can be seen as a task consisting of multiple hierarchies of subtasks with different difficulty levels.

To this end, we propose a concept of Hierarchical Primitive Composition (HPC), where each of the known predefined/pretrained primitives is assembled in a hierarchy to construct an overall compound policy. The main challenge is to properly coordinate simultaneous activation of skills with inconsistent action spaces. Thus, we compose the primitives via multiplicative Gaussians with a factored sum, as is illustrated in Section~\ref{HPC:ActionComposition}. Additionally, we focus on recursively constructing the hierarchies of skills to structuralize our architecture into multiple levels of hierarchy. We test the proposed methodology on the \texttt{pick-and-place} task with a 6 DoF manipulator and examine the effects of simultaneous activation, multi-level hierarchy, and our action composition with ablation studies. The contributions of this work are: 
\begin{itemize}
    \item HPC can simultaneously activate the skills with inconsistent action spaces. This will allow any kinds of pretrained policies to be modularized as primitives and thus increase reusability and modularity.
    \item HPC tackles a complex task through multiple hierarchies in a recursive fashion, which builds on a collection of primitives from other tasks or the HPC module itself.
    \item HPC greatly reduces the learning time of the complicated tasks by abstracting the scope of the action leveraging on the surrogate Markov decision process (MDP).
\end{itemize}
%%%%%%%%%%%%%%%%%%%%%%%%%%%%%%%%%%%%%%%%%%%%%%%%%%%%%%%%%%%%%%%%%%%%%%%%%%%%%%%%

\section{Related Work}
\label{RelatedWork}
Studies on learning reusable skills have long been of an interest and still remain an active research area~\cite{jacobs1991adaptive}. Transfer learning schemes~\cite{devin2017learning,scheiderer2020transfer} involve priorly training an agent on relevant or comparatively easier tasks and fine-tuning the agent on newly appointed tasks.\cite{hasenclever2020comic}~explores this area further to examine how well a skill can be learned and transferred by disentangling the dimension of the problem. While they have successfully reduced the amount of data required, catastrophic forgetting can occur when the gradients of the new task flow through the existing skill primitives~\cite{rusu2016progressive}.

Policy modularization mitigates such incurrence by regarding each primitive as a module and detaching the gradient flow of the new task. \cite{sahni2017learning} learns modules of skill-specific embedder and aggregates the embeddings of all relevant skills with a compositional layer to then learn a policy that maps the composition of embeddings to an action. On the other hand, other approaches involve separating the agent into a hierarchy where the higher level agent aims to properly coordinating the lower level module. One class of studies focuses on sequentially chaining the modules. \cite{frans2017meta, merel2018hierarchical, marzari2021towards}~naively arange the modules to be activated one after another, when~\cite{dalal2021accelerating} provides additional arguments to each primitive as a subgoal with a fixed time horizon. \cite{pore2020simple} also activates one skill at a time using a high-level choreographer, but the skills and the choreographer share the same feature extraction layer. \cite{clegg2018learning, lee2021adversarial}~try to resolve the mismatch between the terminal and initial state distribution of the chained primitives. However, this will restrict the behavior to the prescribed primitive, not allowing the composition or interpolation of multiple primitives to produce new skills. 

Thus, another class of studies involves orchestrating the modules in parallel. \cite{xu2019toward} assemble each attribute from skills by solving the constrained optimization problem, whereas \cite{peng2019mcp} forms a collection of Gaussian policies via multiplications, where each of primitives are independently factored by weights given a state and goals. \cite{qureshi2019composing} also composes a mixture policy of subskills by factoring them with weights from the encoder-decoder model consist of bi-directional recurrent neural networks. 
% This model encodes each action from the skills to the latent states using bi-directional recurrent neural networks and decodes such latent encodings into mixture weights. 
Notwithstanding that the parallel execution allows the primitives to be activated simultaneously, these works restrict the action space of each primitive to be identical to one another. \cite{lee2020learning} incorporates agent-specific primitives and provides skill index with behavior embedding from the meta policy to append more agents, but still lacks a methodology on properly fusing multiple skills of each agent.

We build upon the formulation of \cite{peng2019mcp} and further improve it by structuring multiple levels of hierarchical primitives in a recursive fashion. In contrast to the proposed scheme in~\cite{peng2019mcp}, the role of each of the primitives is known, and the weights distributed over the primitives can display the intention of the overall policy. Moreover, we differentiate the dimensionality of states and actions among primitives and thus, remove the constraints on selecting a proper state and action spaces. This plays an important role when it comes to reusability. We verify this property in the experiments with tasks with and without the gripper angle. Lastly, our novel action composition rule stabilizes the action dimensions that are produced from a single primitive. This is also discussed in the ablation study..
%%%%%%%%%%%%%%%%%%%%%%%%%%%%%%%%%%%%%%%%%%%%%%%%%%%%%%%%%%%%%%%%%%%%%%%%%%%%%%%%

\section{Hierarchical Primitive Composition}
\label{HPC}
% In this section, we describe the framework of the hierarchical primitive composition and its training algorithm. We first define the MDP structure of the unit compound task and expand this concept as a hierarchical task formulation in a recursive manner. We then describe how the compound policy is constructed from the primitives according to the defined MDP and explain the data flow of the model structure. Next, we explain the finalized action generated by exploiting primitive actions and its hierarchical policy design. Lastly, we devise a training algorithm of HPC by reformulating the maximum entropy objectives along with the existing soft-Q and -V functions.

\subsection{Overview}
\label{HPC:Overview}
% Our model consists of a modularized composition of primitives. 
We assume that a single intricate task can be decomposed into a collection of sub-tasks. A primitive is assigned to each sub-task to achieve the goal of the corresponding tasks. These primitives then compose a compound policy with respect to the weights given from a meta-policy, where this compound policy now tries to solve the overall goal of the original task. It is worth noting that the difficulties of tasks are relative, and in some cases, a task once considered to be complex might become a sub-task of even more complex problems. Intuitively, this indicates that a compound policy can also be a primitive of another compound policy. Therefore, we design our structure in a recursively hierarchical fashion and manually assign a \textbf{level} to the compound policy and the corresponding task that indicates how many stacks of modules comprise it. We denote the compound policy, the meta-policy, and the primitive of $i^{th}$ sub-task at level $l$, by $\Pi_l$, $\pi_l^{meta}$, and $\pi_{li}$, respectively. Our goal here is to learn the optimal $\pi_l^{meta}$ given primitives such that that maximizes the policy objective of task $l$.

\subsection{The MDP Formulation}
\label{HPC:MDP}
In this section, we formulate the MDP for task $l$ via MDPs of its sub-tasks to induce the optimal policy $\Pi_l^{*}$. 
We then introduce the meta-MDP, the surrogate of the original MDP. Since the meta-MDP greatly reduces the original action space by mapping compound actions to weights, the scope of the exploration also diminishes and thus increases the sample efficiency.
It will later be proven that the optimal policy of this surrogate MDP, $\pi_l^{meta*}$, is identical to the original optimal policy, $\Pi_l^{*}$.

To formulate the $l$-th MDP for task $l$, we first consider the case for its sub-task $i$. Sub-task $i$ at $l$ is allocated with its own MDP $M_{li}=\{\mathcal{S}_{li}, \mathcal{A}_{li}, \mathcal{P}_{li}, \mathcal{R}_{li}, \gamma\}$ where each element within the tuple shares the same formulation with the standard MDP setting. $M_{li}$ has its own optimal policy, $\pi_{li}$ , which we call a primitive. This primitive can either be a flat policy or another compound policy at level $l-1$, $\Pi_{l-1}$. Please note here that the state and action space of each primitive within the same task level can differ among others.

Once we define MDPs of all sub-tasks within task $l$, we then construct a compound MDP for task $l$ , $M_{l}=\{\mathcal{S}_{l}, \mathcal{A}_{l}, \mathcal{P}_{l}, \mathcal{R}_{l}, \gamma\}$. The state and action spaces of the compound task $l$ is defined to be a jointed space of its primitives: $\mathbf{s}_l \in \mathcal{S}_l = \bigcup_i \mathcal{S}_{li}$ and $\mathbf{a}_l \in \mathcal{A}_l = \bigcup_i \mathcal{A}_{li}$, respectively. The transition probability $\mathcal{P}_l = \mathcal{P}(\mathbf{s}'_l|\mathbf{s}_l,\mathbf{a}_l)$ , the reward function $\mathcal{R}_l=r(\mathbf{s}_l, \mathbf{a}_l)$, and the discount factor $\gamma$ are the same.

To induce the solution for $M_l$ we employ the surrogate MDP of $M_l$, a meta-MDP $M_l^{meta}$ defined by a tuple $\{\mathcal{S}_l^{meta}, \Omega_l, \mathcal{P}_l^{meta}, \mathcal{R}_l^{meta}, \gamma \}$. The state space $\mathcal{S}_l^{meta}$ is identical to the $\mathcal{S}_l$ under the assumption that the same information is required to induce the same solution. 
$[\omega_{l1},\cdots, \omega_{li},\cdots] = \boldsymbol\omega_l \in \Omega_l$ denotes a collection of weights which represents the measure of the influence of each primitive $\pi_{li}$ on $\Pi_l$. Since the users have the knowledge on to which task the primitives are designated, the weight indicates the purpose of the current action. The weight is the output from the meta-policy $\pi_l^{meta}$ given state. Although the weight $\boldsymbol\omega_l$ theoretically needs not be bounded \cite{peng2019mcp}, we restrict its lower bound to be 0, and the sum of all weights to be 1 in order to regard its elements as parameters for a categorical distribution. We later employ the entropy of the meta-policy as the entropy of this distribution to define the maximum entropy objective of the meta-MDP.
Weights $\omega_l$ are analogous to the actions of conventional MDPs since they lead to the next state. The only difference from the original setup is that the weights are no longer stochastic but are deterministically chosen from each state. 
$\mathcal{P}_l^{meta}=\mathcal{P}(\mathbf{s}'_l|\mathbf{s}_l, \boldsymbol\omega_l)$ is a state transition probability of the next state $\mathbf{s}'_l$ given $\mathbf{s}_l$ and $\boldsymbol\omega_l$. $\mathcal{R}_l^{meta} = r^{m}(\mathbf{s}_l, \boldsymbol\omega_l)$ is a reward function of a state-weight pair. We define the reward function $r^{m}(\mathbf{s}_l, \boldsymbol\omega_l)$ to be equal to the expectation of the reward of the compound MDP over actions of $\Pi_l$ given same states and weights (Definition~\ref{def:reward}). The discount factor $\gamma$ is the same.

\subsection{Model Construction}
\label{HPC:ModelConstruction}
% In this section, we describe how the compound policy $\Pi_l$ is constructed, how the data flows, and how the model calculates the primitive actions. The schematic of the model can be found in Fig.~\ref{fig:model_struct}.

\begin{figure}[tp]
    \centering
    \includegraphics[width=\columnwidth]{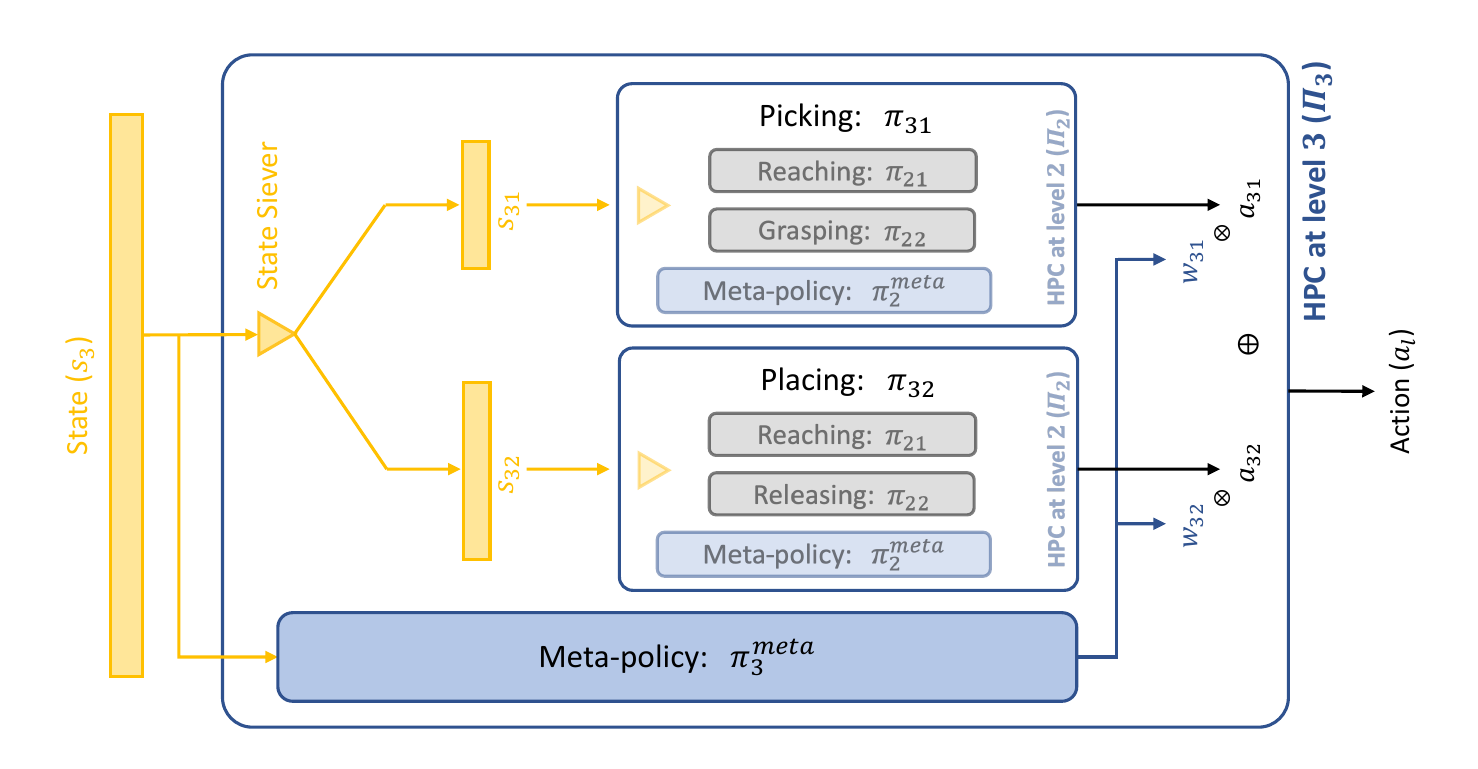}
    \caption{The Structure of HPC.}
    \label{fig:model_struct}
\end{figure}

The compound policy is composed of two key structures: the meta-policy and primitives, as shown in Fig.~\ref{fig:model_struct}. First, the meta-policy receives a state $\textbf{s}_l$ from the space identical to that of the compound MDP and produces weights $\boldsymbol\omega_{l}$. Simultaneously, the same state $\textbf{s}_l$ is also given to the state siever. Since the state space from the compound MDP is the union of all the state spaces of the primitives, this layer sieves $\textbf{s}_l$ into $\textbf{s}_{li}$ and distributes them to the primitives accordingly. Given the states, the corresponding actions are calculated from the primitives. These actions are then combined with the weights to form an overall action of the compound policy.

\subsection{Action Composition}
\label{HPC:ActionComposition}
Each primitive $\pi_{li}(\textbf{a}_{li}|\textbf{s}_{li})$ is regarded as a Gaussian distribution $\mathcal{N}(\boldsymbol\mu_{li}(\textbf{s}_{li}),\Sigma_{li}(\textbf{s}_{li}))$, where $\boldsymbol\mu_{li}(\textbf{s}_{li})\in \mathbb{R}^{|\mathcal{A}_{li}|}$ and  $\Sigma_{li}(\textbf{s}_{li})=$ diag$(\{\sigma_{li,j}^2(\mathbf{s}_{li})\}_{j=1}^{|\mathcal{A}_{li}|})$ denote a mean vector and a diagonal covariance matrix given the state $\mathbf{s}_{li}$, respectively.
We formulate the compound policy at level $l$, $\Pi_l$, to be the multiplication of primitives powered by their respective normalized weights. This is related to the product of experts method in the sense that each skill constrains a different subset of the action space conditioned on the states, while their product constrains the combination of them~\cite{hinton2002training}. We weigh the mean of each element of the action dimension with the summation of the weights of primitives to which they activate on. The probability distribution of the $j$-th action parameter of $\Pi_{l}$, $\Pi_{l,j}$ can be formalized as follows.
    \begin{multline}\label{eq:compound_policy}
        \Pi_{l,j}(a_{l,j}|\mathbf{s}_{l}) = \\
        \frac{1}{Z(\mathbf{s}_{l})} \prod_{i=1}^{P} \mathcal{N}_i\biggl(\mu_{li,j}(\mathbf{s}_{li})\sum_{c \in \mathcal{C}_j} \omega_c(\mathbf{s}_{l}), \sigma_{li,j}^2(\mathbf{s}_{li})\biggr)^{\hat{\omega}_i}
    \end{multline}
where $\mathcal{C}_j$ denotes a set of primitives that include the $j$-th parameter within their action space, $P$ the total number of primitives of task $l$, and $\hat{\omega}_i=\omega_i(\mathbf{s}_{l})/\sum_{p=1}^P \omega_p(\mathbf{s}_{l})$ the normalized weight over the sum. Since the multiplication of Gaussian distributions powered by some factor yields another Gaussian distribution \cite{peng2019mcp}, it allows the resulting policy to be used as another primitive at a higher level. The component-wise parameters for the composed Gaussian distribution can be derived as follows.
    \begin{equation}\label{eq:muandsigma}
        \begin{split}
            \mu_{l,j}&=\frac{\sum_{c \in \mathcal{C}_j} \omega_c}{\sum_{p=1}^P \omega_p/\sigma_{lp,j}^2} \sum_{i=1}^P \frac{\omega_i}{\sigma_{li,j}^2}\mu_{li,j}, 
            \\
            \sigma_{l,j}^2 &= \frac{\sum_{c\in\mathcal{C}_j}\omega_c}{\sum_{p=1}^P \omega_p/\sigma_{lp,j}^2}
        \end{split}
    \end{equation}
where $\mu_{l,j}, \sigma_{l,j}, \omega_c, \omega_p, \omega_i$ are functions of $\mathbf{s}_l$ and $\mu_{li,j}, \sigma_{li,j}, \sigma_{lp,j}$ are functions of $\mathbf{s}_{li}$. Intuitively, the mean of the compound policy indicates the sum of mean of each primitive factored by their weights over variance, and re-weighted by the summed weights which are relevant to the respective action dimension. The compound variance is the reciprocal of the summation of weights over variances and again, re-weighted by the relevant weights. 

\subsection{Training Procedure}
\label{HPC:Training}
The training of HPC is sequentially done from level 2 and iteratively stacks up the learned policies as the level increases. The primitives at level 1 are flat since no compositions occur below that level. To train the compound policy at level $l$, we assume that it is accessible to all primitives that consist of the compound policy. Although which primitive to be used is manually determined by the user, the required engineering effort is minimal as the meta-policy learns how to take advantage of the primitives regardless of their role. This can be verified in Section~\ref{Experiment:TaskPerformance}.

For the meta-policy optimization, any RL methods that are capable of maximizing the meta-objective function defined in Equation~\ref{eq:objective_meta} can be applied, while we used Soft-Actor-Critic (SAC) in this paper~\cite{haarnoja2018soft}. This can easily be done by regarding weights as an action. The parameters of the primitives are kept fixed to prevent optimization instabilities~\cite{gu2016q} and invalid off-policy samples~\cite{nachum2018data}.
%%%%%%%%%%%%%%%%%%%%%%%%%%%%%%%%%%%%%%%%%%%%%%%%%%%%%%%%%%%%%%%%%%%%%%%%%%%%%%%%

\section{Optimal Policy Equivalence}
\label{Algorithm}
In this section, we first describe the maximum entropy objective of our MDP setting at task $l$, and show that the objective of the meta-MDP is identical to that of the original MDP. Then, we prove that the solution for the meta-MDP applies to the standard MDP as well under several assumptions.

\label{Algorithm:OptEquiv}
The objective of reinforcement learning corresponds to the accumulated sum of discounted rewards as follows \cite{haarnoja2018acquiring}.
    \begin{multline*}
        J(\pi)=\sum_{t=0}^{\infty}\mathbb{E}_{(\mathbf{s}_t,\mathbf{a}_t)\sim\rho_\pi}\Biggl[\sum_{k=t}^\infty\gamma^{k-t}\mathbb{E}_{\mathbf{s}_k\sim \mathcal{P},\mathbf{a}_k\sim\pi}[r(\mathbf{s}_k,\mathbf{a}_k)]\Biggr]
    \end{multline*}
To formulate the objective of our MDP, we substitute the standard policy $\pi$ with the compound policy, $\Pi$. In addition, to incentivize the uncertainty of the meta-policy, we introduce the entropy of the primitive weights and formulate it based on a maximum entropy formulation. We omit the subscript $l$ hereafter for simplicity and rewrite the objective as follows.
    \begin{multline}\label{eq:objective_compound}
        J(\Pi)=\sum_{t=0}^{\infty}\mathbb{E}_{(\mathbf{s}_t,\mathbf{a}_t)\sim\rho_\Pi}\Biggl[\sum_{k=t}^\infty\gamma^{k-t}\mathbb{E}_{\mathbf{s}_k\sim \mathcal{P}, \mathbf{a}_k\sim\Pi}[r(\mathbf{s}_k,\mathbf{a}_k) \\
        +\alpha\mathcal{H}(\boldsymbol\omega_k)|\mathbf{s}_t,\mathbf{a}_t]\Biggr]
    \end{multline}
where $H(\boldsymbol\omega)$ denotes the entropy of the categorical distribution with $\boldsymbol\omega$ as its parameter. The entropy term desires the case where the weights are equally distributed to the primitives, and not biased.

Prior to express \eqref{eq:objective_compound} with the meta-policy, we first define the reward function of the meta-MDP as discussed in Section~\ref{HPC:MDP}.
    \begin{definition}[Meta Reward Function]\label{def:reward}
    For any state $\mathbf{s}\in \mathcal{S}$, weight $\boldsymbol\omega = \pi^{meta}(\mathbf{s}) \in \Omega$, and the compound policy defined by the weights $\Pi(\boldsymbol\omega)$, the reward function of the meta-MDP is defined by
        \begin{equation}
            r^{m}(\mathbf{s},\boldsymbol\omega) \coloneqq \mathbb{E}_{\mathbf{a}\sim\Pi(\boldsymbol\omega)}[r(\mathbf{s},\mathbf{a})|\mathbf{s}]
        \end{equation}
    \end{definition}
We next define the objective of the meta-MDP as follows.
    \begin{definition}[Meta Objective Function]\label{def:objective_meta}
    The objective of the meta-MDP given a meta-policy is defined by
        \begin{multline} \label{eq:objective_meta}
            J(\pi^{meta})=\sum_{t=0}^{\infty}\mathbb{E}_{\mathbf{s}_t\sim\mathcal{P}}\Biggl[\sum_{k=t}^\infty\gamma^{k-t}\mathbb{E}_{\mathbf{s}_k\sim\mathcal{P}}[r^{m}(\mathbf{s}_k,\boldsymbol\omega_k)\\+\alpha\mathcal{H}(\boldsymbol\omega_k)|\mathbf{s}_t]\Biggr]
        \end{multline}
    where, $\boldsymbol\omega_k = \pi^{meta}(\mathbf{s}_k)$
    \end{definition}

From the above definitions, we now can show that the objective of the compound policy is equivalent to that of the meta-policy.
    \begin{restatable}[Objective Function Equivalence]{lemma}{primelemma}\label{lem:equiv}
        From Definitions~\ref{def:reward} and~\ref{def:objective_meta}, we obtain the following equivalence.
        \begin{equation}
            J(\pi^{meta}) = J(\Pi)
        \end{equation}
    \end{restatable}
    \begin{proof}
    See Appendix~\ref{appa}.
    \end{proof}

We assume that there exists an optimal compound policy which behaves identical to the optimal standard policy as follows.
    \begin{assumption}\label{ass:compound_optimality}
    For any MDPs, there exists an optimal compound policy $\Pi^*$ such that
        \begin{equation}
            J(\Pi^*) = J(\pi^*)
        \end{equation}
    where $\pi^*$ is the optimal standard policy.
    \end{assumption}
This assumption guarantees that once we find the compound policy which optimizes the compound policy objective function, it is the optimal solution of the MDP, $\pi^*$.

Finally, it can be straightforwardly shown that the below satisfies.
    \begin{theorem}[Optimal Policy Equivalence]\label{thm:optimal_equivalence}
    From Lemma~\ref{lem:equiv} and Assumption~\ref{ass:compound_optimality}, for the optimal meta-policy $\pi^{meta*}$ which optimizes the objective \eqref{eq:objective_meta},
        \begin{equation}
            \pi^{meta*} = \pi^*
        \end{equation}
    \end{theorem}
    \begin{proof}
    From Lemma~\ref{lem:equiv}, the objective of the optimal meta-policy and the optimal compound policy is the same, thus from Assumption~\ref{ass:compound_optimality}, the following holds.
        \begin{equation}
            J(\pi^{meta*}) = J(\Pi^*) = J(\pi^*)
        \end{equation}
    \end{proof}
From Theorem~\ref{thm:optimal_equivalence}, it becomes possible to devise an algorithm which seeks for the optimal meta-policy instead of the standard policy.
%%%%%%%%%%%%%%%%%%%%%%%%%%%%%%%%%%%%%%%%%%%%%%%%%%%%%%%%%%%%%%%%%%%%%%%%%%%%%%%%

\begin{figure*}[h]
    \centering
    \includegraphics[width=\textwidth]{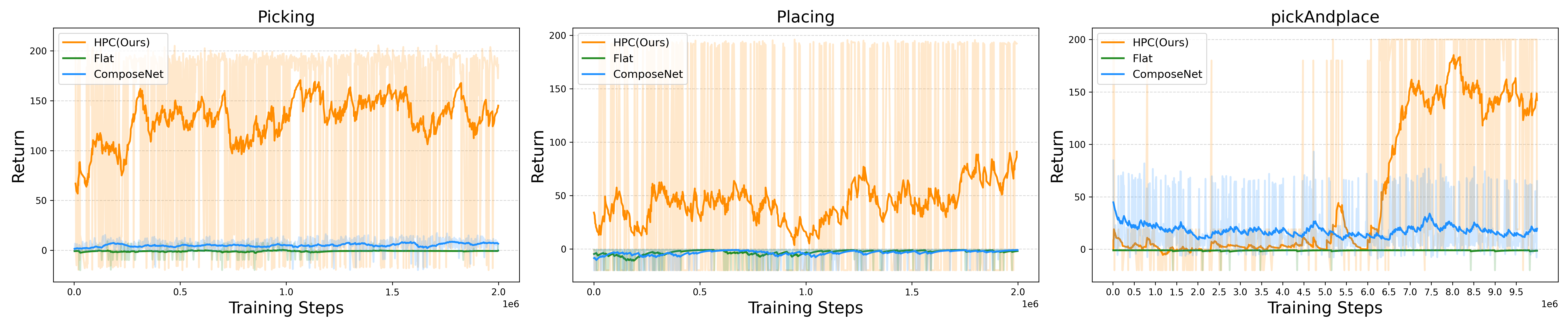}
    \caption{Training curves of each tasks. The x-axis shows the number of training steps and y-axis represents the episode return.}
    \label{fig:lc_task}
\end{figure*}

\section{Experiment}
\label{Experiment}
\subsection{Experimental Setup}
\label{Experiment:Setup}

As shown in Fig.~\ref{fig:task_schematic}, we have newly designed the environment for the following six skills composing \texttt{pick-and-place}. The descriptions of each tasks are as follows:
(1) \textbf{\texttt{Reaching}}: The end-effector moves from the initial pose to the target pose. 
(2) \textbf{\texttt{Grasping}}: To grasp the target object and lift it above 0.1m. Since \texttt{grasping} is intended to focus solely on grasping the object, we bound the initial position of the end-effector to be within 0.15m to the object. 
(3) \textbf{\texttt{Releasing}}: To release the object at the destination. Since \texttt{releasing} is intended to focus solely on releasing the object, the initial pose is sampled near the destination with the object in hand. 
(4) \textbf{\texttt{Picking}}: Identical to \texttt{grasping} except for its initial pose, as it starts farther away from the object. 
(5) \textbf{\texttt{Placing}}: Identical to \texttt{releasing} except for its initial pose, as it starts farther away from the object. 
(6) \textbf{\texttt{Pick-and-place}}: To pick up an object and place it at the given destination.

To validate the effectiveness of HPC on composing skills with different action dimensions, we differentiate the dimensionality of \texttt{releasing} with the others by assigning 6 dimensions (incremental pose of the end-effector), while others include gripper angle as an additional component of the controllable action. We refer the readers to Appendix~\ref{appb} for the details regarding state dimensions of each skill.

All the environments are created within the MuJoCo physics simulator \cite{todorov2012mujoco}. For the manipulator, the 6-DoF Jaco2 model from Kinova Robotics is used. To control the end-effector, we used an operational space controller (OSC) within the abr\_control package provided from Applied Brain Research \cite{khatib1987unified, DeWolf2017}. OSC receives the target pose of the end-effector and calculates required torques to be applied to each joint of the manipulator. 

\subsection{Results and Analysis}
\label{Experiment:Results}

To examine the effects of HRL when solving the complex task, we compare HPC to the vanilla SAC denoted as Flat \cite{haarnoja2018soft}. To evaluate the effects of pretrained primitives in terms of the performance, 
we compare HPC to ComposeNet~\cite{sahni2017learning}. To apply ComposeNet to our setting, we detached the last layer of the primitives from our experiment and used the remaining networks as skill trunks. We then stacked a compositional layer along with a policy layer on top of the fixed skill trunks. Shared policy layers are not used.

\subsubsection{\textbf{Training}}
\label{Experiment:Training}
The left and the center plots from Fig.~\ref{fig:lc_task} show the training curves of both level 2 tasks - \texttt{picking} and \texttt{placing}, respectively. The results show that HPC outperforms the baselines by far, while both Flat and ComposeNet do not experience a successful episode. The right plot from Fig.~\ref{fig:lc_task} shows the training curve of \texttt{pick-and-place} task. It shows that HPC outperforms both baselines in the level 3 task as well. Since the baselines did not manage to solve the lower-level tasks, it is evident that the level 3 task is also impossible to be solved. Although ComposeNet leverages the pretrained skill trunks to obtain skill embeddings, it still requires its policy and the value function to be optimized for the original action space and thus, suffers from the curse of dimensionality. HPC, however, abstracted its scope of action from the \{pose, gripper angle\} selection to the weight selection problem and managed to solve the task.

\subsubsection{\textbf{Task performance}}
\label{Experiment:TaskPerformance}
\begin{figure*}[!h]\centering
    \includegraphics[width=0.78\textwidth]{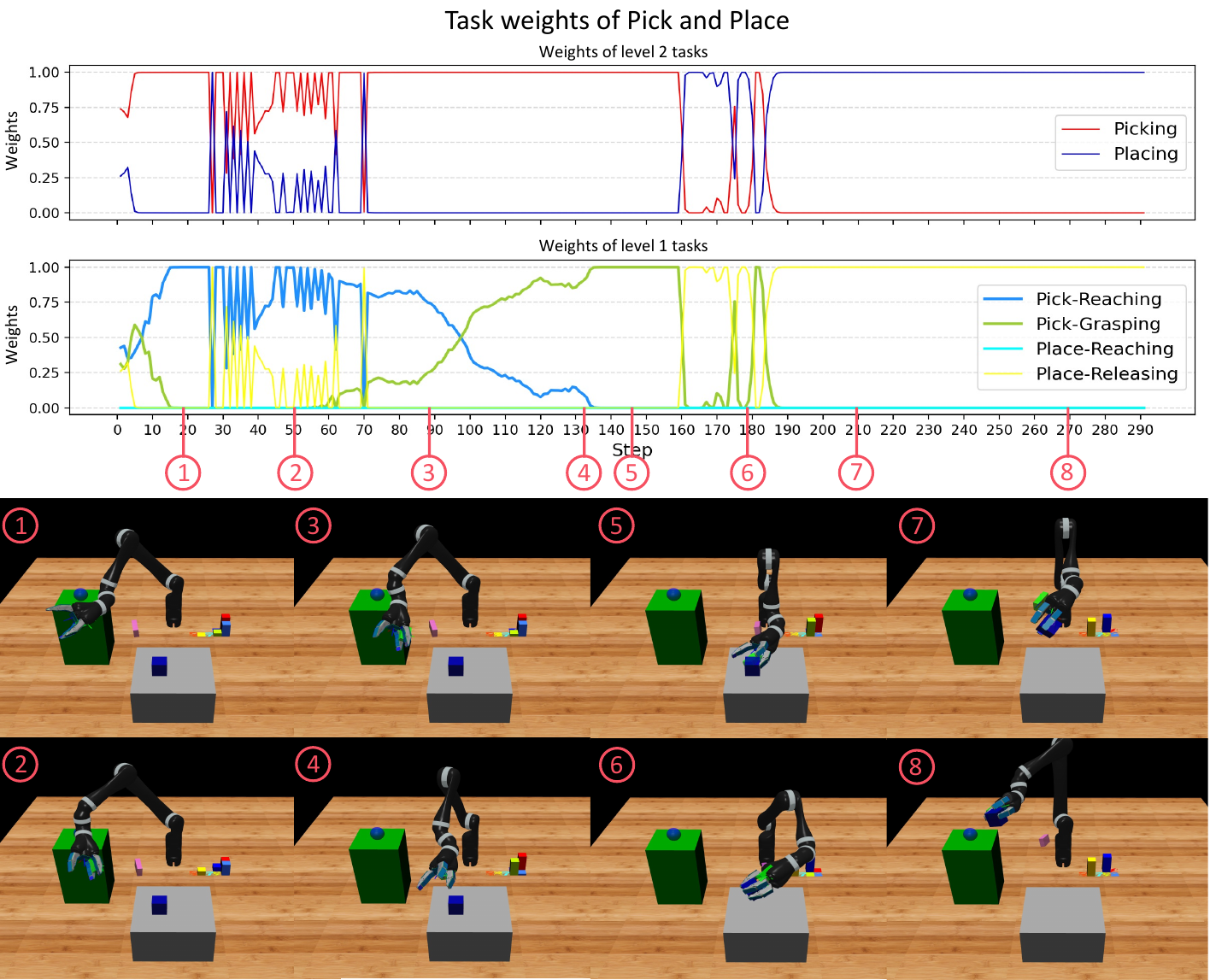}
    \caption{Weight examination of the pick and place task. \textbf{Top:} Weigth plot of the level 2 and level 1 tasks, respectively. \textbf{Bottom:} The visualization of the simulation at the corresponding timestep. The video of the pick and place task can be found in \url{https://youtu.be/_MlKP80sVtE}.}
    \label{fig:weight_sim_task3_pickAndplace}
\end{figure*}

We further examine the performance of HPC with a few frames from \texttt{pick-and-place} task, as shown in Fig.~\ref{fig:weight_sim_task3_pickAndplace}.
The weights distributed to each primitive of level 2 and level 1 are plotted in the upper and the lower part of the weight plot, respectively. In the beginning, the reaching primitive is mainly used to drive the end-effector towards the object (\circled{1}). While approaching, the placing primitive is used to properly orient the end-effector (\circled{2}). This addresses that the meta-policy takes advantage of the primitives within the skill set regardless of their role. It is also worth emphasizing that the weight is not binarily biased to either one of the primitives but rather to take advantage of both. Once the end-effector is well oriented, the picking primitive is fully used until the end-effector successfully grasps the object and lifts it above the table (\circled{3} - \circled{5}). With the object in the gripper, the meta-policy now leverages the placing primitive and completes the overall task (\circled{6} - \circled{8}).

\subsection{Ablation Study}
\label{Experiment:Ablation}

    \begin{figure}[ht]\centering
        \includegraphics[width=\columnwidth]{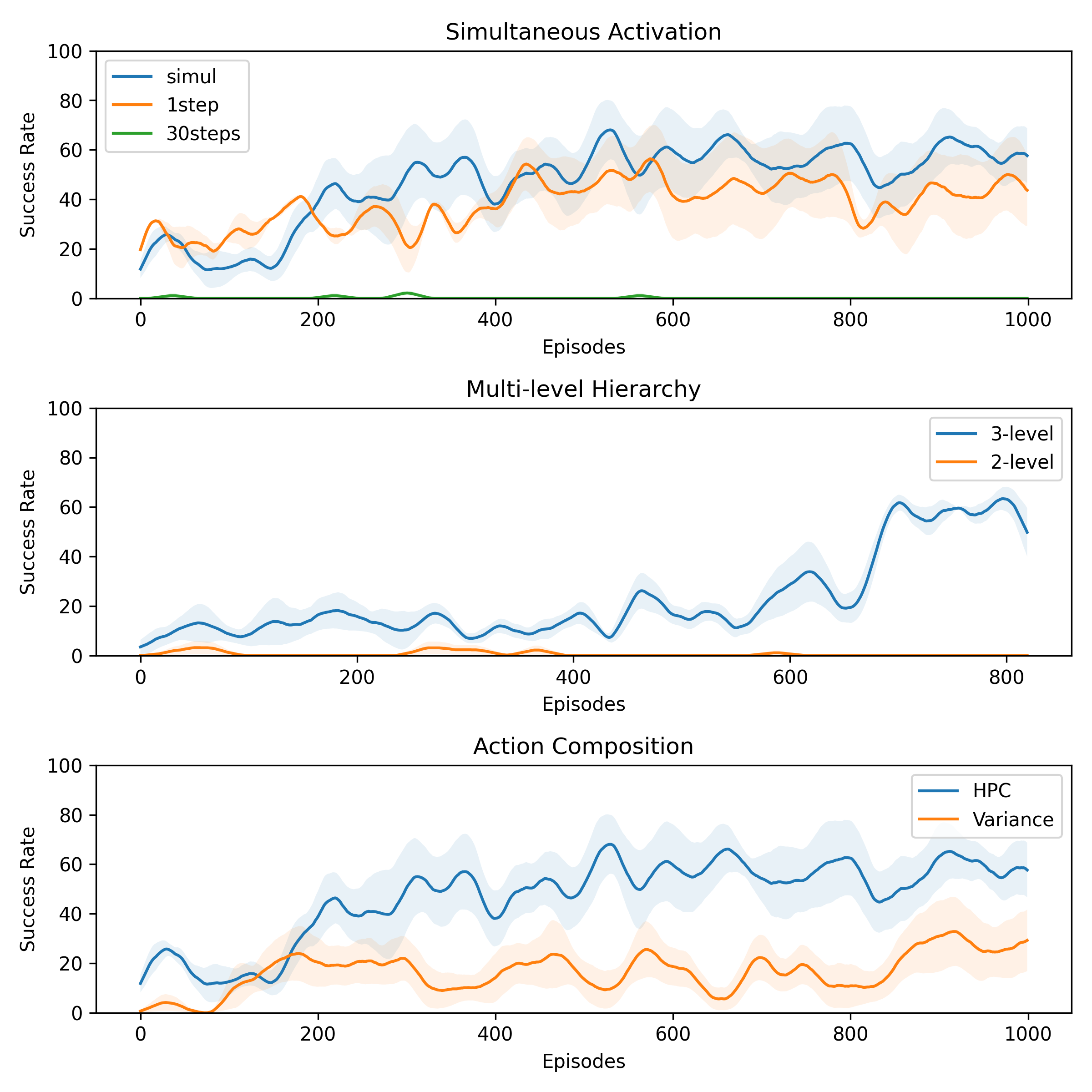}
        \caption{Success rate of the ablation studies. Top plot shows the effect of the simultaneous activation, plot in the middle shows the effect of the multi-level hierarchy, and the bottom plot shows the effect of our action composition.}
        \label{fig:ablation}
    \end{figure}

Our next experiment investigates the effects of simultaneous activation, multi-level hierarchy and our action composition.
\subsubsection{\textbf{Simultaneous activation}}
We first compare the performance of the conventional HPC on \texttt{picking} task with the binary selection of skills with the fixed time horizon of 1 and 30 steps. The former is analogous to the work of~\cite{goyal2019reinforcement,frans2017meta}, where the primitives are either competing or selected to be activated at a given timestep, whereas the latter is to the work of~\cite{dalal2021accelerating} where the higher-level agent assigns which skill to be selected along with the action parameters. From the top plot of Fig.~\ref{fig:ablation}, it can be shown that the simultaneous activation outperforms both of the binary selection setups.
\subsubsection{\textbf{Multi-level hierarchy}}
The following study analyzes the effect of the hierarchy of multiple levels by composing \texttt{pick-and-place} task with level 1 primitives. As the plot in the middle of Fig.~\ref{fig:ablation} represents, adding more levels to the complex task makes the learning efficient. This phenomenon is also in relation to curriculum learning, as the intermediate levels work as stepping stones to guide the agent to seek the beneficial path.
\subsubsection{\textbf{Action composition}}
Lastly, we verify the effect of our action composition rule by comparing the performance of \texttt{picking} task with anoother composition method~\cite{peng2019mcp}. Since the reaching skill does not optimize the gripper-angle, it is possible to add an additional action dimension with high variance to the learned Gaussian. From the bottom plot of Fig.~\ref{fig:ablation}, it can be seen that the proposed action composition method outperforms the opponent. Given high variance, this is due to the gripper angle of the compound action being identical to that of the grasping primitive when computed according to the formulation from~\cite{peng2019mcp}. When the policy encounters the state region in which the grasping primitive was not trained, the action becomes unstable.

\subsection{Failed case: Operational State Space Mismatch}
\label{Experiment:Failure}
\begin{figure}[!b]\centering
        \subfloat[The walking primitive.]{
            \includegraphics[width=0.48\columnwidth]{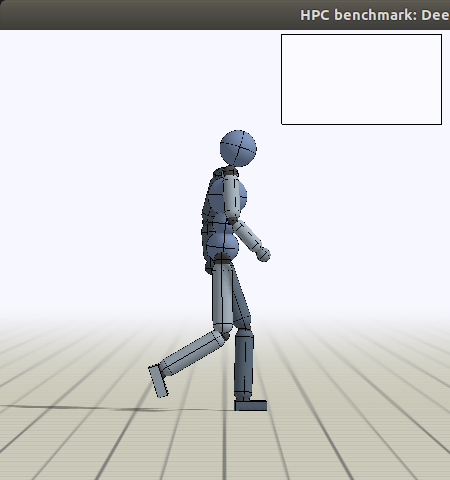}%
            \label{fig:bm_walk}}
        \hfil
        \subfloat[The running primitive.]{
            \includegraphics[width=0.48\columnwidth]{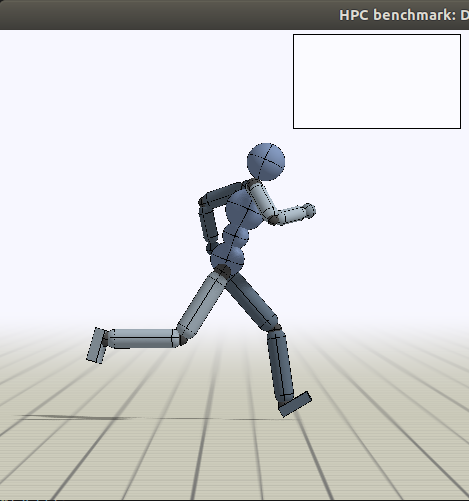}%
            \label{fig:bm_run}}
        \caption{The walking and running primitives of the humanoid robot.}
        \label{fig:bm}
    \end{figure}

In this experiment, we show how HPC can fail to learn the task if the primitives do not share the operational state Space space. With the priorly trained primitive for the humanoid robot walking and running, as shown in Fig.~\ref{fig:bm}, we tried to train the robot to jog at a velocity between walking and running by composing two primitives. Although both primitives share the positional state space of the joint motor, the operational state space of the joint velocity differed. Moreover, the region of the velocity space required for the jogging is between the two primitives, and the meta-policy was not able to seek the optimal action. This will be discussed in the following section.
%%%%%%%%%%%%%%%%%%%%%%%%%%%%%%%%%%%%%%%%%%%%%%%%%%%%%%%%%%%%%%%%%%%%%%%%%%%%%%%%

\section{Discussion and Conclusion}
\label{Conclusion}
In this paper, a concept of hierarchical primitive composition is proposed, where each of the known predefined/pretrained primitives is assembled in a hierarchy to construct a compound overall policy. HPC dramatically reduces the learning time of the complicated robotic tasks by imploying the surrogate MDP. It can represent the intention of the agent by weights distributed to each of the primitives in a human-interpretable manner while reusing pretrained primitives to enhance sample efficiency and guarantee stability in the optimization process. In addition, since HPC is not restricted to the dimensionality of the state and action spaces, any other works on RL can be applied to compose its primitive with pretrained policies. 
% Lastly, \cite{simon1991architecture,meunier2009hierarchical} proposes that the human brain functional networks have a hierarchical modular organization, where medial occipital, lateral occipital, central, parieto-frontal, and fronto-temporal region of the brain takes each of the intrinsic roles. HPC adapts the proposed formulation of the human brain in terms of the hierarchical structure, and thus, other works on recognition and logic using neural networks can be fused into HPC and further structuralize the total artificial intelligence as a whole.

There still remain future works to do. Firstly, Assumption~\ref{ass:compound_optimality} limits the applicability of HPC to the case where the optimal policy is in the set of the compound policies. If there are not enough skills within the set of primitives, the assumption does not hold. Therefore, it is required to employ a newly trainable auxiliary primitive, which closes the gap between the compound policy and the optimal policy. Second, if the primitives do not share the same region of the operational state state space, the compound policy should bias its weight signal on a single primitive activated at the moment to output a proper action. This can be observed from the experiment discussed in Section~\ref{Experiment:Failure}. This can be mitigated with the work from~\cite{lee2021adversarial} to properly match the operational state space of primitives.
%%%%%%%%%%%%%%%%%%%%%%%%%%%%%%%%%%%%%%%%%%%%%%%%%%%%%%%%%%%%%%%%%%%%%%%%%%%%%%%%

\appendices
\section{Proof of Objective Function Equivalence}
\label{appa}
\primelemma*
\begin{proof}
    We first rewrite the compound objective as follows.
    \begin{multline}\label{eq:objective_compound_app}
        J(\Pi)=\sum_{t=0}^{\infty}\mathbb{E}_{(\mathbf{s}_t,\mathbf{a}_t)\sim\rho_\Pi}\Biggl[\sum_{k=t}^\infty\gamma^{k-t}\mathbb{E}_{\mathbf{s}_k\sim \mathcal{P}, \mathbf{a}_k\sim\Pi}[r(\mathbf{s}_k,\mathbf{a}_k)\\
        +\alpha\mathcal{H}(\boldsymbol\omega_k)|\mathbf{s}_t,\mathbf{a}_t]\Biggr]
    \end{multline}

    % By linearity of expectiation, the underlined term becomes
    By the linearity of the expectation,~\eqref{eq:objective_compound_app} becomes
    \begin{multline}\label{eq:underlined}
            \sum_{t=0}^{\infty}\mathbb{E}_{(\mathbf{s}_t,\mathbf{a}_t)\sim\rho_\Pi}\Biggl[\sum_{k=t}^\infty\gamma^{k-t}\mathbb{E}_{\mathbf{s}_k\sim\mathcal{P},\mathbf{a}_k\sim\Pi}[r(\mathbf{s}_k,\mathbf{a}_k)|\mathbf{s}_t,\mathbf{a}_t] \\
            + \mathbb{E}_{\mathbf{s}_k\sim\mathcal{P}}[\alpha\mathcal{H}(\boldsymbol\omega_k)|\mathbf{s}_t]\Biggr]
    \end{multline}

    Note here that the entropy of a categorical distribution of the weights $\mathcal{H}(\boldsymbol\omega_k)$ is independent to the actions, and thus the expectation over the action can be omited. From Definition~\ref{def:reward},~\eqref{eq:underlined} now becomes
    \begin{multline*}
        \sum_{t=0}^{\infty}\mathbb{E}_{(\mathbf{s}_t,\mathbf{a}_t)}\Biggl[\sum_{k=t}^\infty\gamma^{k-t}\mathbb{E}_{\mathbf{s}_k}[r^{m}(\mathbf{s}_k,\boldsymbol\omega_k)|\mathbf{s}_t]\\
        + \mathbb{E}_{\mathbf{s}_k}[\alpha\mathcal{H}(\boldsymbol\omega_k)|\mathbf{s}_t]\Biggr]
    \end{multline*}
    \begin{multline*}
        = \sum_{t=0}^{\infty}\mathbb{E}_{(\mathbf{s}_t,\mathbf{a}_t)}\Biggl[\sum_{k=t}^\infty\gamma^{k-t}\mathbb{E}_{\mathbf{s}_k}[r^{m}(\mathbf{s}_k,\boldsymbol\omega_k)
        +\alpha\mathcal{H}(\boldsymbol\omega_k)|\mathbf{s}_t]\Biggr]
    \end{multline*}
    \begin{multline*}
        = \sum_{t=0}^{\infty}\mathbb{E}_{\mathbf{s}_t}\Biggl[\sum_{k=t}^\infty\gamma^{k-t}\mathbb{E}_{\mathbf{s}_k}[r^{m}(\mathbf{s}_k,\boldsymbol\omega_k)
        +\alpha\mathcal{H}(\boldsymbol\omega_k)|\mathbf{s}_t]\Biggr]
    \end{multline*}
    \begin{equation}\label{eq:exp_tometa}
        = J(\pi^{meta})
    \end{equation}
\end{proof}

\section{State Dimensions}
\label{appb}
\begin{table}[H]
    \caption{States of each task.}
    \begin{center}
        \begin{tabularx}{\columnwidth}{c|Y}
            \toprule
            Task & States \\
            \midrule
            Reaching & $\textbf{p}_{E}$, $\textbf{q}_{E}$, $\textbf{p}_{r}$, $\textbf{q}_{r}$ \\
            Grasping & $i_{E}$, $\textbf{p}_{E}$, $\textbf{q}_{E}$, $\theta_{g}$, $\textbf{p}_{obj}$ \\
            Releasing & $i_{E}$, $\textbf{p}_{E}$, $\textbf{q}_{E}$, $\theta_{g}$, $\textbf{p}_{obj}$, $\textbf{p}_{d}$ \\
            Picking & $i_{E}$, $\textbf{p}_{E}$, $\textbf{q}_{E}$, $\theta_{g}$, $\textbf{p}_{obj}$, $\textbf{p}_{r}$, $\textbf{q}_{r}$ \\
            Placing & $i_{E}$, $\textbf{p}_{E}$, $\textbf{q}_{E}$, $\theta_{g}$, $\textbf{p}_{obj}$, $\textbf{p}_{d}$, $\textbf{p}_{r}$, $\textbf{q}_{r}$ \\
            Pick-and-place & $i_{E}$, $\textbf{p}_{E}$, $\textbf{q}_{E}$, $\theta_{g}$, $\textbf{p}_{obj}$, $\textbf{p}_{d}$, $\textbf{p}_{r}$, $\textbf{q}_{r}$ \\
            \bottomrule
        \end{tabularx}    
    \end{center}
    \label{tab:state_reach}
\end{table}
\noindent
where\\
$i_{E} \in \{0,1\}$ : The touch indicator \\
$\textbf{p}_{E} \in \realR^3$ : The position of the end-effector (\si{\meter}) \\
$\textbf{q}_{E} \in \realR^3$ : The orientation of the end-effector (rad) \\
$\theta_{g} \in \realR$ : The finger angle of the gripper (rad) \\
$\textbf{p}_{obj} \in \realR^3$ : The position of the target object (\si{\meter}) \\
$\textbf{p}_{d} \in \realR^3$ : The orientation of the destination (rad)\\
$\textbf{p}_{r} \in \realR^3$ : The position of the reaching target (\si{\meter})\\
$\textbf{q}_{r} \in \realR^3$ : The orientation of the reaching target (rad)\\
%%%%%%%%%%%%%%%%%%%%%%%%%%%%%%%%%%%%%%%%%%%%%%%%%%%%%%%%%%%%%%%%%%%%%%%%%%%%%%%%

\bibliographystyle{IEEEtran.bst}
\bibliography{bibliography.bib}
%%%%%%%%%%%%%%%%%%%%%%%%%%%%%%%%%%%%%%%%%%%%%%%%%%%%%%%%%%%%%%%%%%%%%%%%%%%%%%%%

\end{document}